\documentclass{article}
\usepackage{spconf,amsmath,amssymb,graphicx,booktabs,xcolor,placeins}
\graphicspath{{figures/}}
\title{GRACE: Geometry- and Ray-Aware Camera-Efficient Multi-View Pedestrian Tracking}

\twoauthors
 {Taigo Sakai, Kazuhiro Hotta}
	{Meijo university\\
    Department of Science Technology\\
    1-501, Shiogamaguchi,\\
    Tempaku, Nagoya 468-8501, Japan}
 {Hiroki Kouno, Naoki Kato}
	{Chubu Electric Power Co., Inc.\\
    1-1 Higashishin-cho, Higashi-ku,\\
    Nagoya 461-8680, Japan}

\begin{document}
\ninept
\maketitle

\begin{abstract}
Reducing the number of cameras lowers deployment cost but leaves fewer views to correct projection errors and temporary detection failures in bird's-eye view (BEV) tracking. Homography projection can spread pedestrian features along the camera viewing direction, while temporary score drops can interrupt an existing track. We introduce GRACE, a camera-efficient multi-view tracker that combines Volumetric-Guided Fusion (VGF) and Ray Conditioning to reduce directional projection errors, and BEV Track Recovery (BTR) to maintain existing tracks through temporary score drops without allowing weak detections to start new tracks. VGF balances homography-projected and 3D-lifted features according to camera coverage, while Ray Conditioning explicitly supplies each camera's viewing direction. With only two cameras, GRACE improves MOTA from 83.54 to 91.07 on WildTrack and from 68.27 to 79.36 on MultiviewX, with the largest BTR gain observed in the two-camera setting.
\end{abstract}

\begin{keywords}
multi-view pedestrian tracking, bird's-eye view, few-camera tracking, volumetric fusion, ray conditioning
\end{keywords}

\section{Introduction}

Multi-view trackers aggregate calibrated camera features in BEV, allowing overlapping views to correct occlusion and localization errors \cite{hou2020mvdet,teepe2024earlybird,teepe2024lifting}. Fewer cameras reduce installation and communication costs, but 
a BEV response stretched away from a pedestrian's true position is then less likely to be corrected by another view.

Two failures dominate this setting. First, homography maps features from a pedestrian's body onto the ground plane, stretching the BEV response along the viewing direction. 
We call it a directional projection error. Second, coverage changes near a field-of-view boundary can lower the detection score for several frames and split one track into two identities. The second failure depends on whether a detection continues an existing track or starts a new one.

GRACE addresses the two failures using our three components: Volumetric-Guided Fusion (VGF), Ray Conditioning, and BEV Track Recovery (BTR). VGF balances homography-projected and volumetrically lifted features using camera coverage. The coverage shows how many cameras observe each ground-plane location but does not show the camera viewing directions. Ray Conditioning supplies the viewing direction at each ground-plane location in the BEV grid.
BTR lets a low-confidence detection continue an existing track. The same detection cannot start a new track.

\begin{figure*}[t]
\centering
\includegraphics[width=0.9\textwidth]{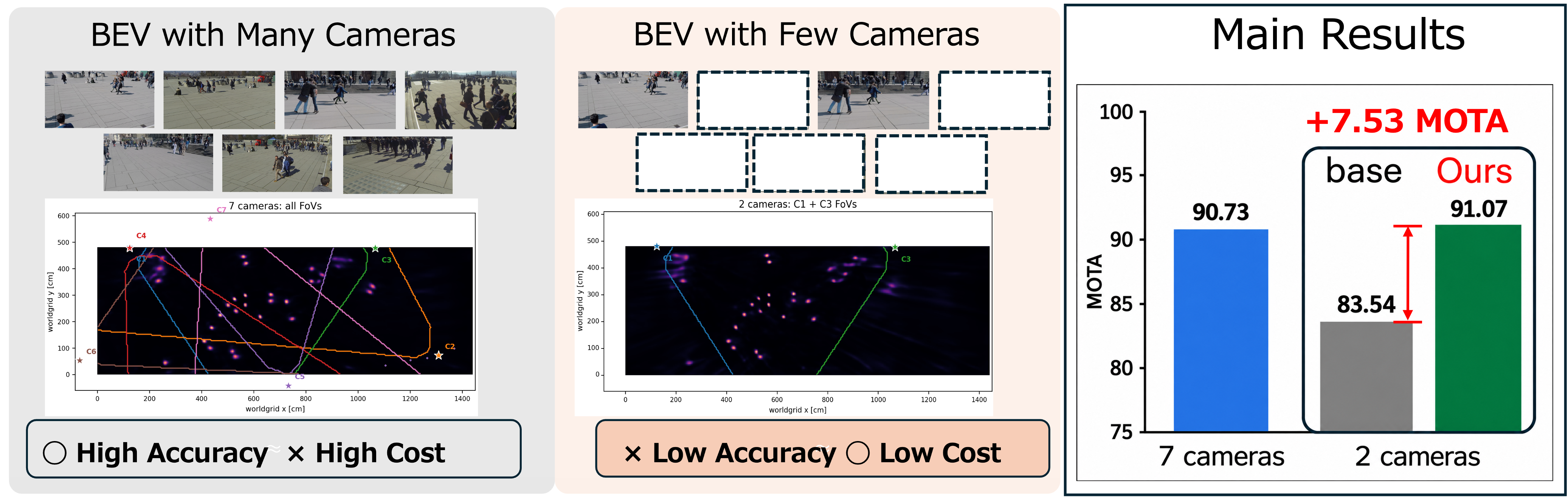}
\caption{Few-camera tracking on WildTrack. GRACE reaches 91.07 MOTA with only two cameras, compared with 83.54 for two-camera TrackTacular. The seven-camera reference uses VGF + Ray detections with the TrackTacular tracker and reaches 90.73 MOTA.}
\label{fig_teaser}
\end{figure*}

TrackTacular is our primary tracking baseline because it also performs detection and tracking in BEV. Figure~\ref{fig_teaser} shows the resulting camera--accuracy trade-off: GRACE reaches 91.07 MOTA with only two WildTrack cameras, compared with 83.54 for two-camera TrackTacular. For reference, the seven-camera setting with VGF + Ray detections and the TrackTacular tracker reaches 90.73 MOTA.

Our contributions are as follows.
\begin{itemize}
    \item We identify directional projection errors and track splits caused by short score drops as two failure modes in few-camera BEV tracking.
    \item We introduce geometry-aware fusion and track recovery that use only calibration and past observations.
    \item We improve two-camera tracking on WildTrack and MultiviewX, reaching 91.07 MOTA on WildTrack and 79.36 MOTA on MultiviewX.
\end{itemize}


\section{The proposed method: GRACE}

\begin{figure*}[t]
\centering
\includegraphics[width=0.92\textwidth]{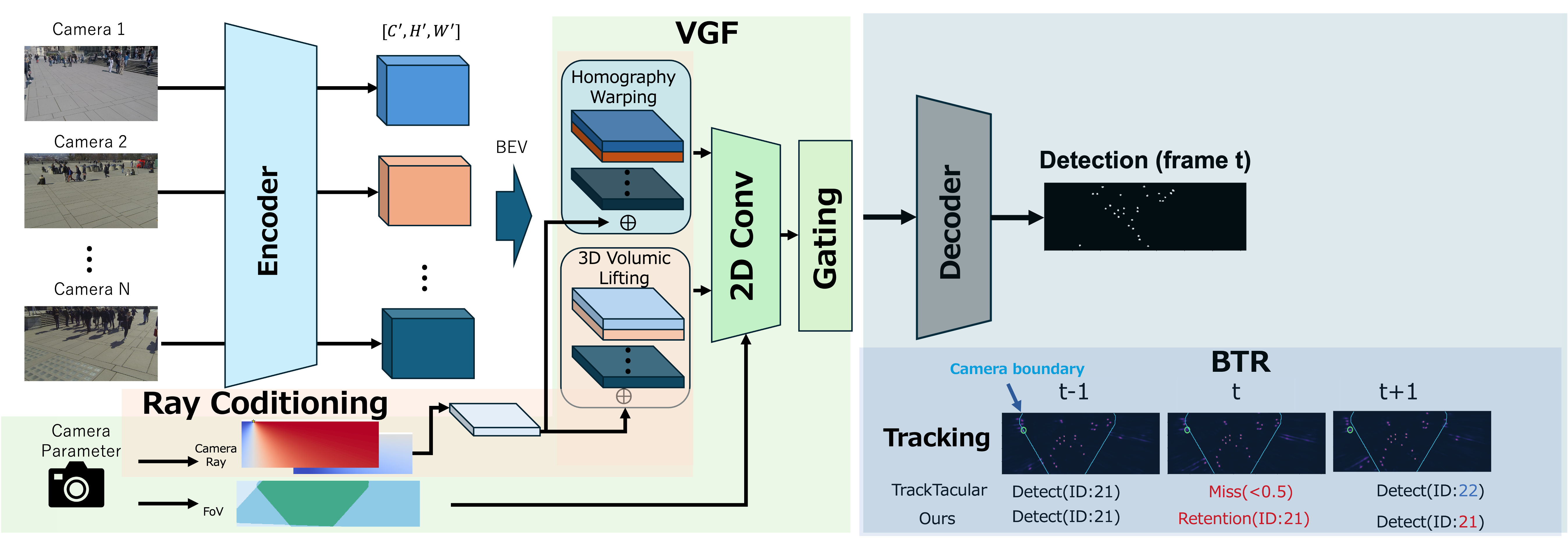}
\caption{GRACE overview. VGF fuses projected and lifted features using coverage and ray geometry. BTR lets low-confidence detections continue existing tracks through short score drops.}
\label{fig_method}
\end{figure*}

Figure~\ref{fig_method} summarizes GRACE. VGF and Ray Conditioning improve BEV fusion from synchronized calibrated views. BTR preserves tracks through short detection gaps.

\subsection{Problem Formulation}

For $K$ synchronized calibrated cameras, camera $c$ provides image $x_t^{(c)}$ at frame $t$. The detector transforms all views to a common BEV and predicts a center heatmap
\begin{equation}
P_t\in[0,1]^{Y\times X},
\end{equation}

With many cameras, projected features from different viewing directions can compensate for a localization error in one view. With few cameras, fewer views contribute to each ground-plane location, so projection errors are more likely to remain after fusion. This is the first failure. Second, a temporary drop in detection confidence can break an existing track and assign a new identity when the pedestrian is detected again. VGF and Ray Conditioning address the first failure, while BTR addresses the second.

\subsection{Volumetric-Guided Fusion}

Homography projects a pedestrian's body features onto the ground plane, which can spread the projected activation away from the true foot position along the viewing direction. Figure~\ref{fig_bev_camera_count}(a) illustrates this effect conceptually, while Fig.~\ref{fig_bev_camera_count}(b) and (c) show the difference between two- and seven-camera TrackTacular outputs.
Volumetric projection instead lifts image features through calibrated 3D space before reducing them to BEV.

Let $H_t$ be the homography-projected feature and $V_t$ the volumetrically lifted feature. VGF also derives the coverage map $m$ from calibration.
\begin{equation}
m\in[0,1]^{Y\times X}.
\end{equation}
We define $g$ as the per-location mixing weight between the volumetric feature $V_t$ and the homography-projected feature $H_t$. A larger $g$ gives more weight to $V_t$.
\begin{align}
g_{\mathrm{learn}}
&=
\sigma\!\left(
\mathrm{Conv}_{1\times1}
\left(
\operatorname{concat}(H_t,V_t,m)
\right)
\right),\\
g
&=
\tfrac12 g_{\mathrm{learn}}
+
\tfrac12(1-m),\\
Z_t
&=
g\odot V_t+(1-g)\odot H_t .
\label{eq_vgf}
\end{align}

The coverage prior increases the volumetric weight where few cameras provide mutual correction. In overlapping regions, multiple homography features can reinforce a correctly localized response, so the learned gate can retain more of $H_t$. The learned term also adapts this balance to the observed features rather than using coverage alone.

\subsection{Ray Conditioning}

\begin{figure}[t]
\centering

\includegraphics[width=0.9\columnwidth]{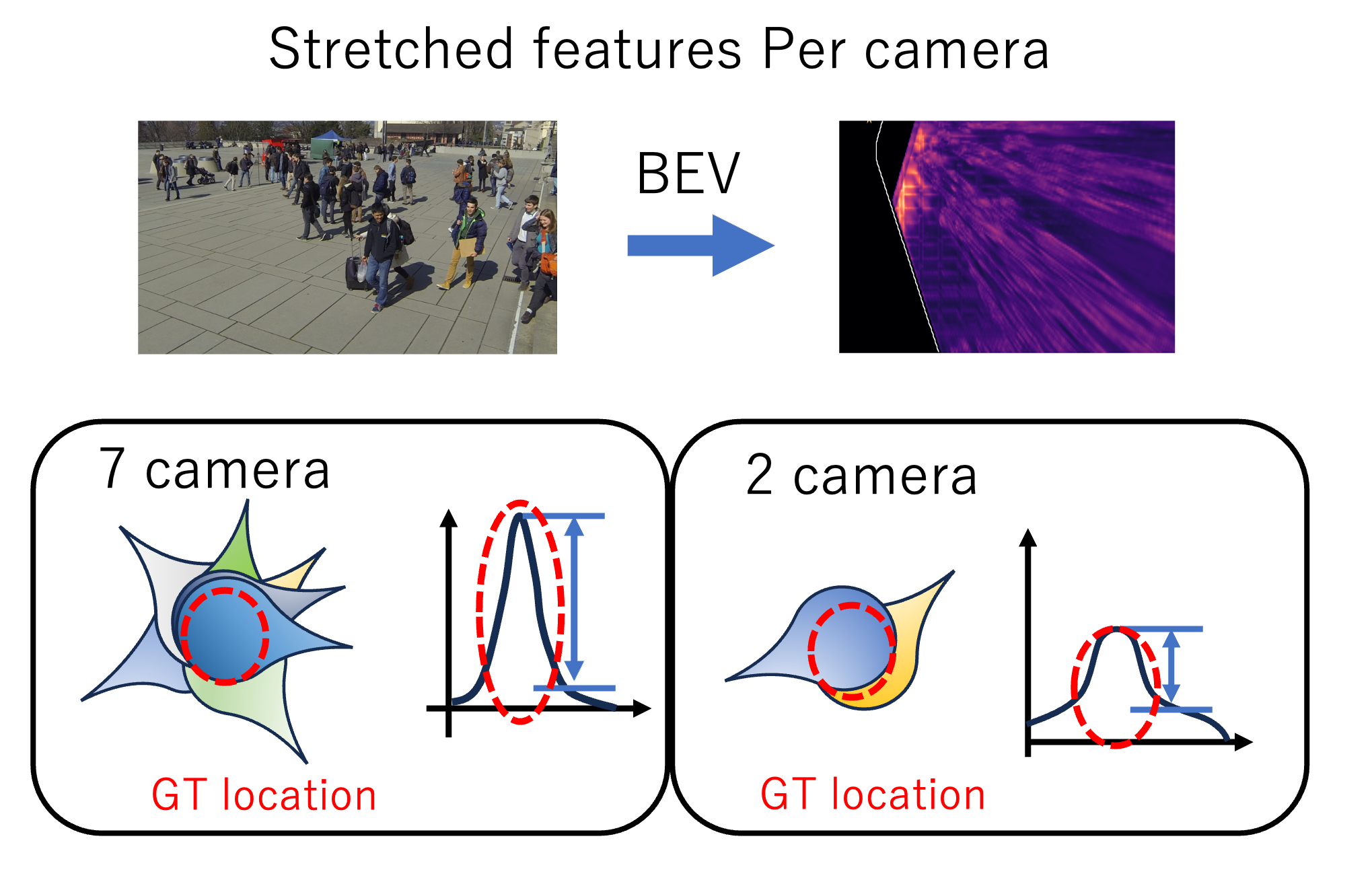}\\[-1mm]
{\small (a) Conceptual illustration}

\vspace{1mm}

\begin{minipage}{0.38\columnwidth}
    \centering
    \includegraphics[width=\linewidth]{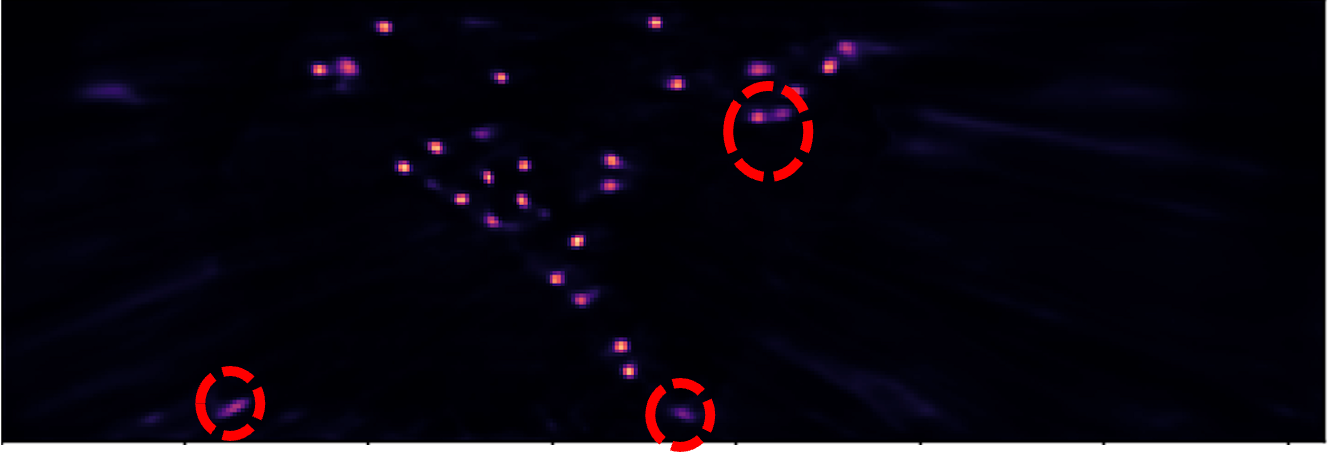}\\[-1mm]
    {\small (b) TrackTacular with 2 cameras}
\end{minipage}
\hfill
\begin{minipage}{0.38\columnwidth}
    \centering
    \includegraphics[width=\linewidth]{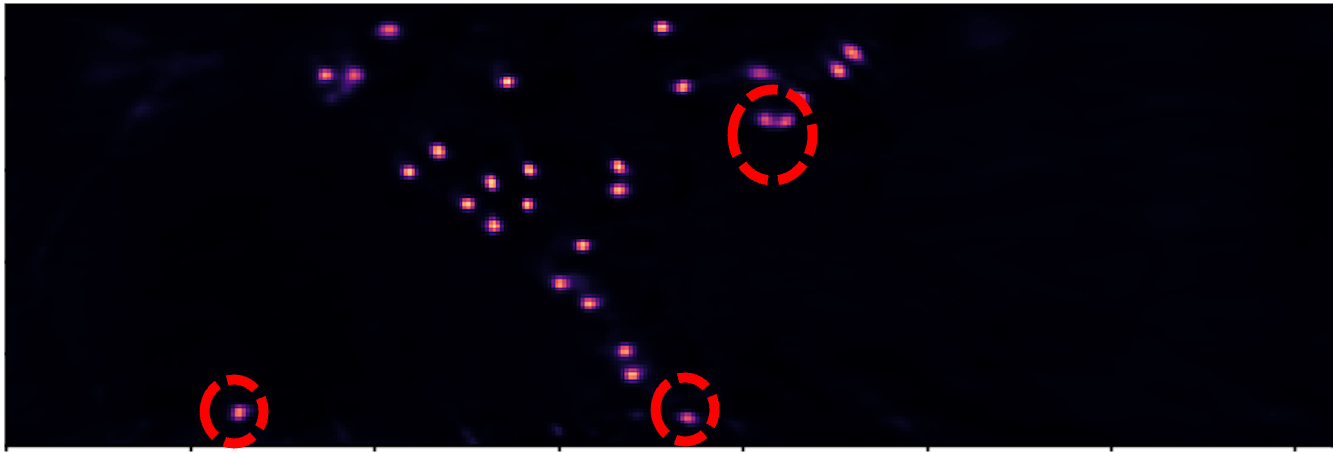}\\[-1mm]
    {\small (c) TrackTacular with 7 cameras}
\end{minipage}

\caption{Effect of camera count on BEV projection. Directional spreading is more visible with two cameras, while additional views produce more concentrated BEV responses.}

\label{fig_bev_camera_count}
\end{figure}

Figure~\ref{fig_bev_camera_count} illustrates why viewing direction becomes important with few cameras. The conceptual example shows that projection can stretch pedestrian responses along the camera viewing direction. The actual TrackTacular BEV maps show the same tendency. With two cameras, directional spreading remains more visible, while additional views produce more concentrated responses.

This motivates Ray Conditioning. We explicitly provide the viewing direction of each camera during BEV fusion.
Several multi-camera 3D detectors
encode the 3D position, camera viewpoint, or ray geometry to relate image features to 3D space \cite{liu2022petr,xiong2023cape,shu2023p3d,chen2023vedet,chu2024rayformer,liu2024raydn}.

Coverage does not identify the direction along which a projection error spreads. For the center $\mathbf q_{ij}$ of ground-plane cell $(i,j)$ in the BEV grid and calibrated camera position $\mathbf o^{(c)}$, we define
\begin{equation}
\mathbf r^{(c)}_{ij}
=
\frac{\mathbf q_{ij}-\mathbf o^{(c)}}
{\|\mathbf q_{ij}-\mathbf o^{(c)}\|_2}
=
\left(r_x^{(c)},r_y^{(c)}\right).
\label{eq_ray}
\end{equation}

We collect $\mathbf r_{ij}^{(c)}$ over all BEV cells to form a fixed two-channel ray map
$R^{(c)}\in\mathbb{R}^{2\times Y\times X}$.
Its two channels contain $r_x^{(c)}$ and $r_y^{(c)}$ at each ground-plane location.

The fixed two-channel map $R^{(c)}$ 
is computed once from calibration and concatenated with camera $c$'s volumetric feature. 
The two maps visualize the horizontal and vertical components of the viewing direction at each ground-plane location.

\begin{equation}
\widetilde U_t^{(c)}
=
\rho\!\left(
\operatorname{concat}\left(U_t^{(c)},R^{(c)}\right)
\right),
\end{equation}
The compressed features are fused as
\begin{equation}
V_t^{\mathrm{ray}}
=
\psi_{\mathrm{vol}}
\left(
\{\widetilde U_t^{(c)}\}_{c}
\right),
\label{eq_raycond}
\end{equation}
where $\rho$ compresses each camera feature and $\psi_{\mathrm{vol}}$ fuses cameras. We substitute $V_t^{\mathrm{ray}}$ for $V_t$ in Eq.~\ref{eq_vgf}. Unlike the scalar coverage map, $R^{(c)}$ distinguishes locations observed from different directions even when their camera counts match. Camera positions and BEV coordinates already provide the required supervision-free signal.

\subsection{BEV Track Recovery}

Online MOT predicts and associates objects over time \cite{zhou2020centertrack}. Representative online MOT methods include ByteTrack~\cite{zhang2022bytetrack}
and SGT~\cite{hyun2023sgt}. ByteTrack shows that low-score detections can still be useful when associating an existing track.

Near a field-of-view boundary, the detection score may fall below threshold for one or two frames. BTR retains the standard threshold $\tau=0.5$ for new tracks, but matches existing tracks to detections down to $\tau-\delta$ with $\delta=0.2$ using the normal association rule. A low-confidence detection cannot initialize a track, limiting false identities. If no detection exists, the motion model predicts for at most $L=2$ frames when the previous confidence is at least $0.4$. The track terminates if matching still fails. BTR remains online and never uses future observations.


\section{Experimental Setup}

\begin{figure}[t]
\centering
\includegraphics[width=0.38\columnwidth]{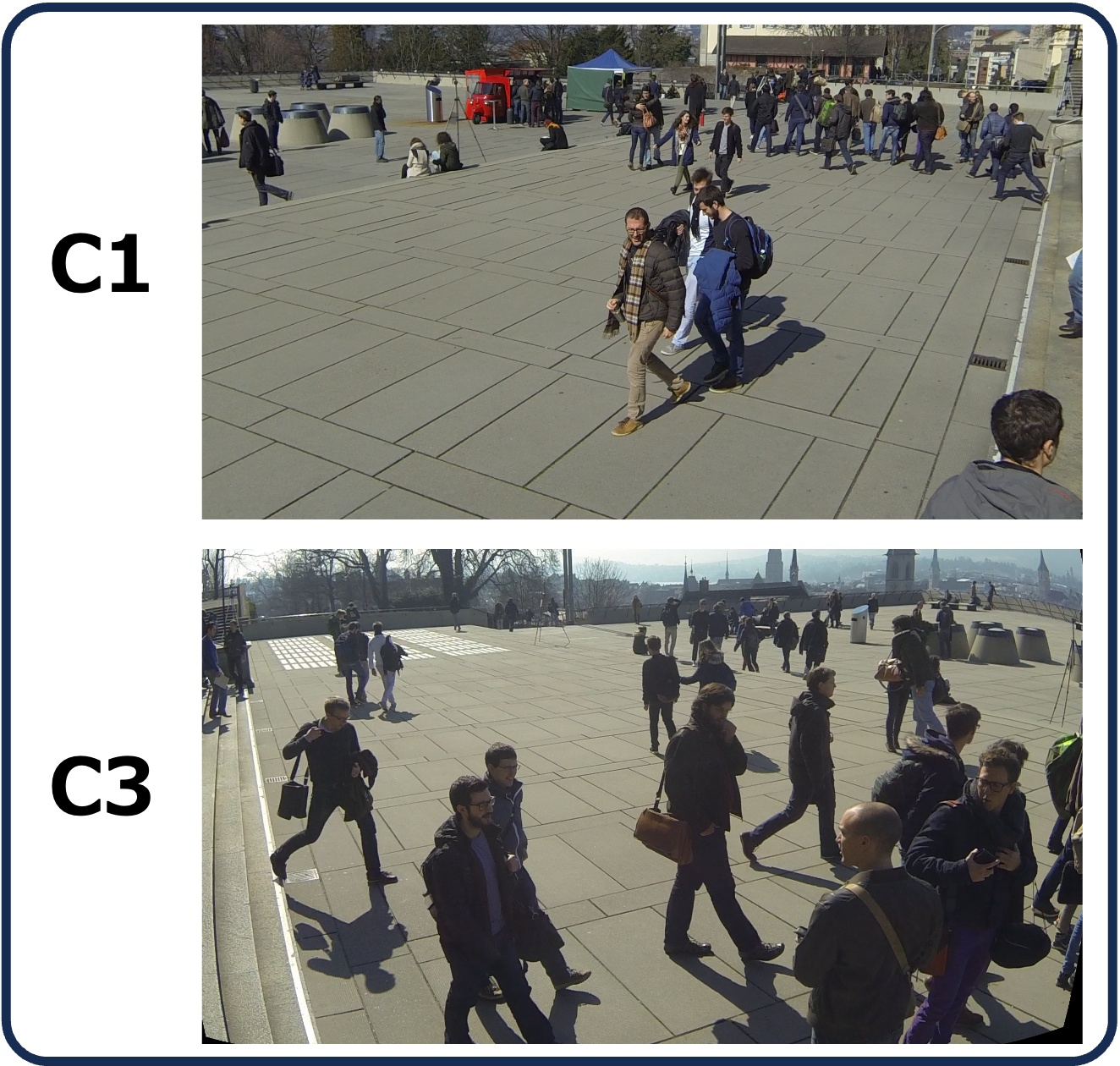}
\hfill
\includegraphics[width=0.38\columnwidth]{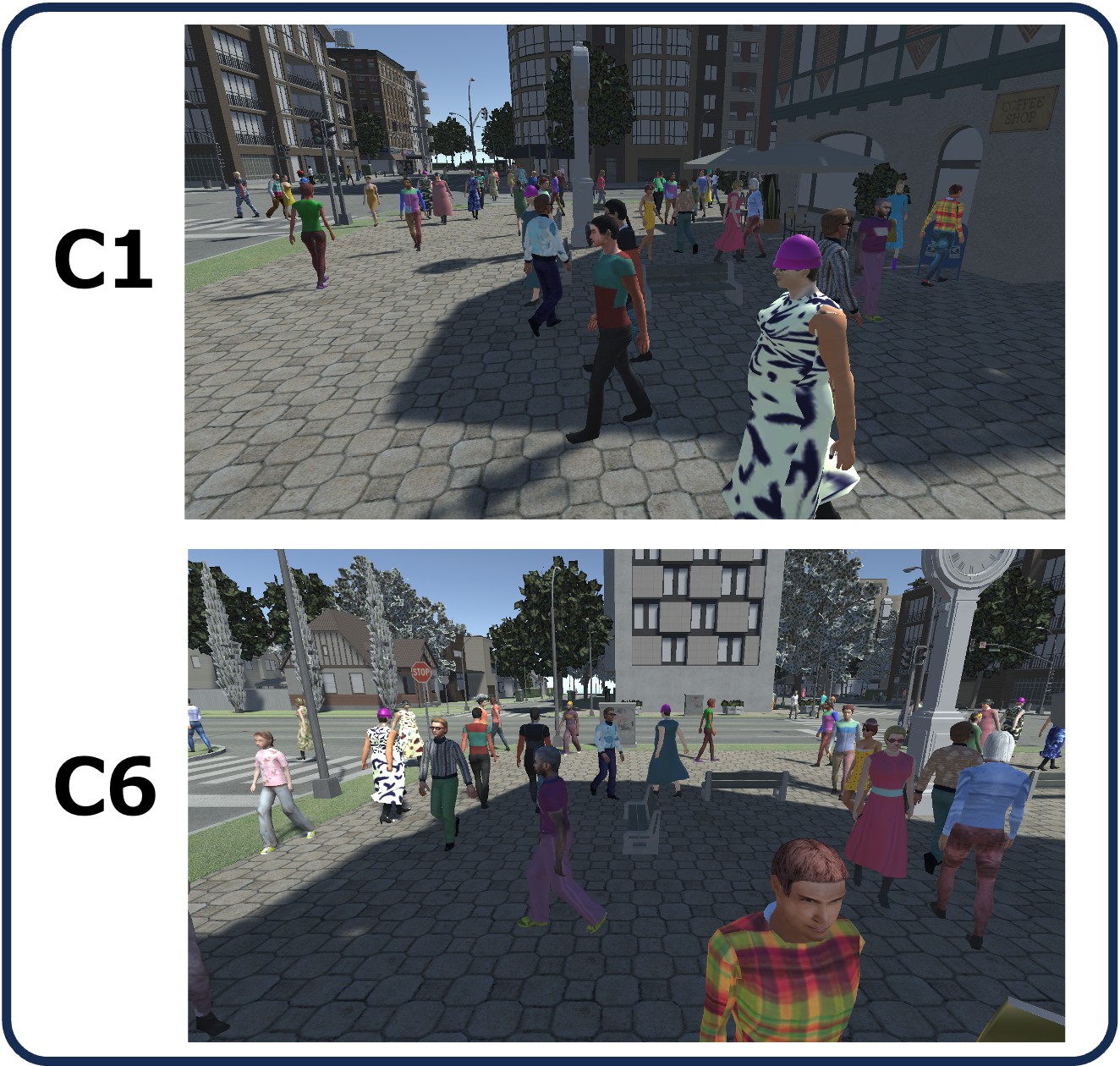}
\caption{Sample images from the two-camera settings. WildTrack C1+C3 is shown left and MultiviewX C1+C6 right.}
\label{fig_dataset_views}
\end{figure}

We evaluate on the standard BEV regions of WildTrack \cite{chavdarova2018wildtrack} and MultiviewX \cite{hou2020mvdet}. All methods use the same camera pair on each dataset. Figure~\ref{fig_dataset_views} shows sample images from these two-camera settings: C1+C3 for WildTrack and C1+C6 for MultiviewX.

\subsection{Camera Configuration}

\begin{figure}[!t]
\centering
\includegraphics[width=0.9\columnwidth]{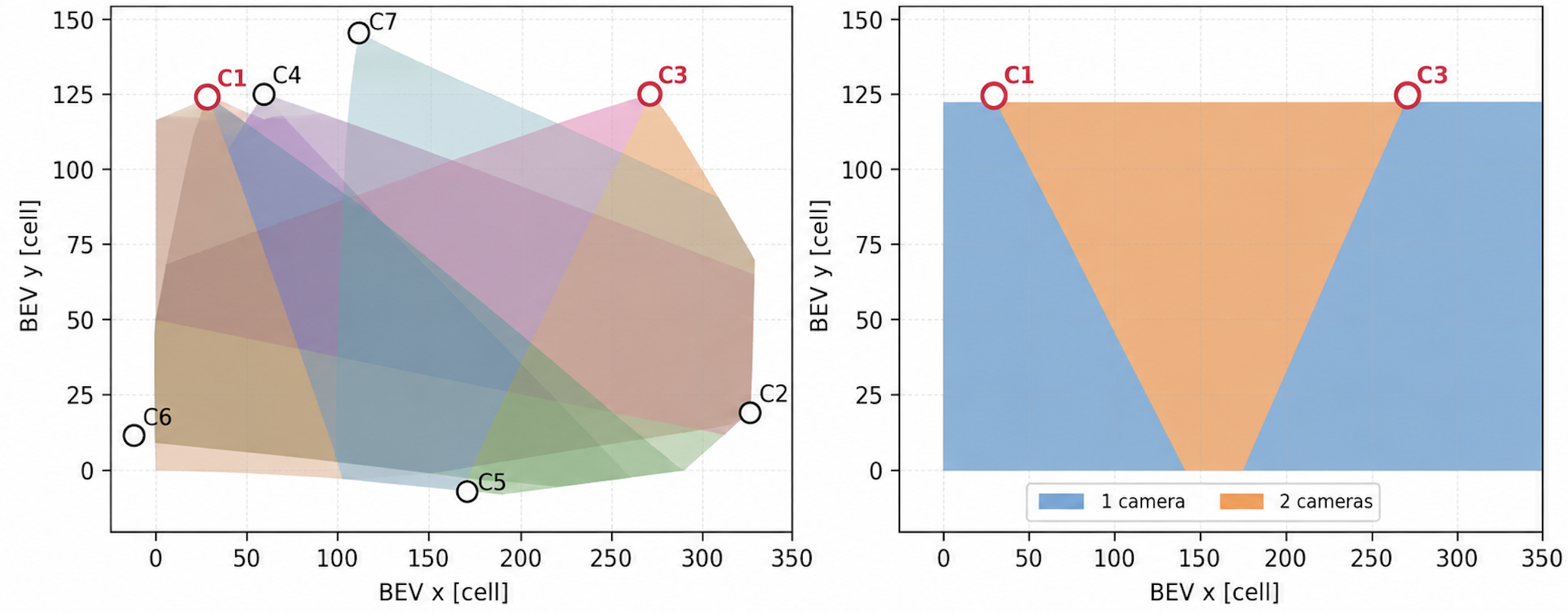}
\caption{WildTrack camera configuration. Left column shows the positions and fields of view of all seven cameras. Right column shows the selected C1+C3 pair. Blue regions are observed by one camera and the orange region by both cameras. C1+C3 covers the complete BEV evaluation area, with $44.98\%$ observed by both cameras.}
\label{fig_camera_configuration}
\end{figure}

We focus on the two-camera setting because it is the minimum configuration that still enables multi-view fusion and represents the most challenging camera-efficient setting. We also evaluate three to seven cameras in the camera-count study to examine how the effect of GRACE changes as more views become available. The two-camera pair is selected
using only camera calibration before evaluating any model. 
Our first criterion is complete coverage of the BEV evaluation area. Among all WildTrack two-camera pairs, C1+C3 is the only pair that satisfies this requirement, and $44.98\%$ of the BEV grid is visible from both cameras. This provides full scene coverage while retaining a substantial overlap for multi-view fusion. Figure~\ref{fig_camera_configuration} shows the resulting configuration. The pair is fixed for all WildTrack two-camera experiments. For MultiviewX, C1+C6 covers $95.62\%$ of the complete BEV evaluation area with at least one camera, while $44.21\%$ of the whole area is visible from both cameras.

\vspace{-2.5mm}

We follow TrackTacular preprocessing and its ResNet-18 training configuration \cite{teepe2024lifting,he2016resnet}. Methods implemented in our framework share the camera subset, input processing, detection threshold, and evaluator. CaMuViD and RMCS are detection-only comparison methods: CaMuViD predictions are converted to our WildTrack coordinates, and RMCS is trained on the selected pair. We report standard detection and tracking metrics \cite{nascimento2006detection,bernardin2008clear,ristani2016id}.

We select one BTR parameter set on WildTrack C2+C5 and MultiviewX C1+C5. These tuning pairs differ from the reported two-camera pairs to avoid tuning on the reported results. We use the selected parameters unchanged at every reported camera count.
Ray controls replace calibrated rays with zero, constant, or wrong-camera channels while leaving the remaining pipeline unchanged. VGF adds 2.16M parameters, Ray Conditioning 13,856, and BTR none. The full model runs at 12.66 FPS on an RTX A6000, with 0.027 ms BTR overhead.

\section{Results}

\subsection{WildTrack Results}

Table~\ref{tab_main} reports the main WildTrack comparison. With the same C1+C3 input, GRACE improves TrackTacular from 82.53 to 86.38 MODA and from 83.54 to \textbf{91.07 MOTA}. CaMuViD and RMCS are detection-only methods, so only their tracking cells are omitted. The seven-camera block is limited to TrackTacular and GRACE.

\subsection{Camera-Count Study}

\begin{figure}[t]
\centering
\includegraphics[width=\columnwidth]{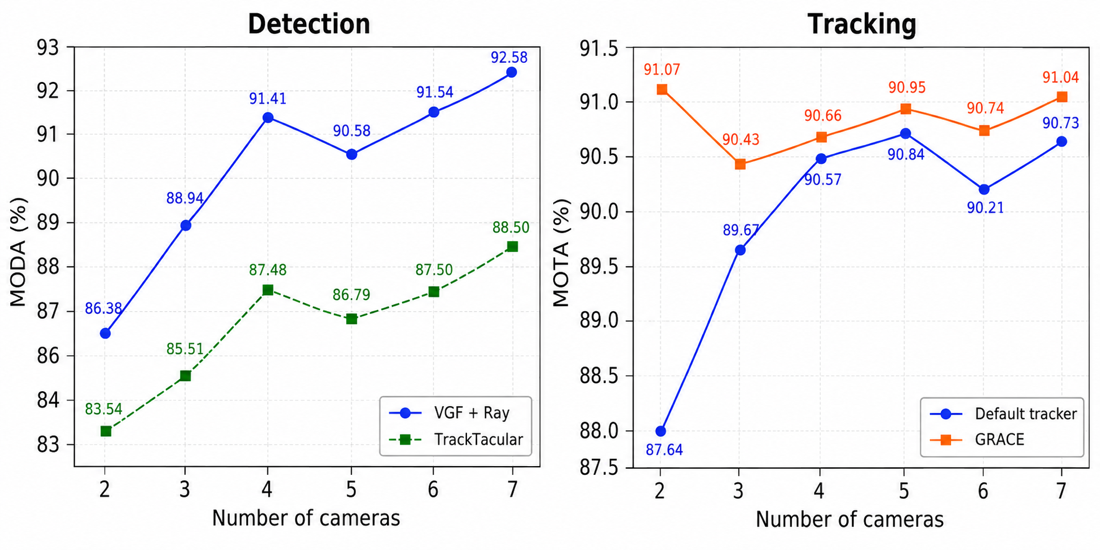}
\caption{WildTrack camera-count study. BTR gives its largest gain with two cameras.}
\label{fig_count}
\end{figure}

Across two to seven WildTrack cameras (Fig.~\ref{fig_count}), VGF + Ray consistently improves detection: MODA rises from 86.38 to 92.58, versus 83.54 to 88.50 for TrackTacular. BTR adds 3.43 MOTA points with two cameras but at most 0.76 with three to seven. This concentration of gain supports its role in recovering gaps that additional views would otherwise cover. Parameters remain fixed.

\begin{table}[t]
\centering
\caption{WildTrack results with two and seven cameras. CaMuViD and RMCS are detection-only methods. Bold: best within each camera count.}
\label{tab_main}
\resizebox{\columnwidth}{!}{%
\begin{tabular}{lcccccccc}
\toprule
Method & Cams & MODA & MODP & Prec. & Recall & IDF1 & MOTA & MOTP\\
\midrule
TrackTacular~\cite{teepe2024lifting} & 2 & 83.54 & 69.53 & 97.93 & 84.31 & 83.81 & 83.54 & 79.73\\
LiftNet~\cite{teepe2024lifting} & 2 & 80.32 & 74.75 & \textbf{98.03} & 81.97 & 87.67 & 82.63 & 83.34\\
CaMuViD~\cite{daryani2025camuvid} & 2 & 58.61 & 66.42 & 79.87 & 78.36 & -- & -- & --\\
RMCS~\cite{li2026rmcs} & 2 & 77.42 & \textbf{78.55} & 94.74 & 81.97 & -- & -- & --\\
\midrule
\textbf{GRACE} & 2 & \textbf{86.38} & 76.86 & 97.76 & \textbf{88.41} & \textbf{94.21} & \textbf{91.07} & \textbf{84.37}\\
\midrule
TrackTacular~\cite{teepe2024lifting} & 7 & 88.50 & 78.56 & \textbf{97.42} & 91.39 & 94.00 & 89.71 & 84.90\\
\textbf{GRACE} & 7 & \textbf{92.58} & \textbf{80.33} & 96.42 & \textbf{96.15} & \textbf{94.53} & \textbf{91.04} & \textbf{86.26}\\
\bottomrule
\end{tabular}}
\end{table}

\subsection{MultiviewX Results}

On MultiviewX (Table~\ref{tab_mvx}), GRACE reaches \textbf{79.36 MOTA} with only two cameras, 11.09 MOTA points above TrackTacular. MODA increases from 71.26 to 77.09 and IDF1 from 67.34 to 75.05. In the full six-camera setting, we compare TrackTacular and GRACE. In the two-camera C1+C6 setting, we evaluate TrackTacular, LiftNet, CaMuViD, and GRACE using the same camera input.

\begin{table}[t]
\centering
\caption{MultiviewX results. Upper: TrackTacular and GRACE with six cameras. Lower: the same two-camera input for all methods. CaMuViD is a detection-only method. Bold: best within each camera count.}
\label{tab_mvx}
\resizebox{\columnwidth}{!}{%
\begin{tabular}{lccccccc}
\toprule
Method & Cams & MODA & MODP & Recall & IDF1 & MOTA & MOTP\\
\midrule
TrackTacular~\cite{teepe2024lifting} & 6 & 93.64 & \textbf{88.03} & 94.65 & 78.08 & \textbf{89.89} & \textbf{83.86}\\
\textbf{GRACE} & 6 & \textbf{95.69} & 87.72 & \textbf{96.50} & \textbf{78.21} & 87.39 & 83.48\\
\midrule
TrackTacular~\cite{teepe2024lifting} & 2 & 71.26 & \textbf{84.09} & 71.71 & 67.34 & 68.27 & \textbf{82.56}\\
LiftNet~\cite{teepe2024lifting} & 2 & 58.77 & 83.40 & 59.42 & 57.50 & 57.39 & 79.43\\
CaMuViD~\cite{daryani2025camuvid} & 2 & 59.70 & 79.93 & \textbf{84.41} & -- & -- & --\\
\midrule
\textbf{GRACE} & 2 & \textbf{77.09} & 78.50 & 77.67 & \textbf{75.05} & \textbf{79.36} & 78.88\\
\bottomrule
\end{tabular}}
\end{table}

\subsection{Component Contributions}

Table~\ref{tab_component} separates the three components on WildTrack C1+C3. Relative to TrackTacular, VGF improves MODA by 2.52, IDF1 by 7.43, and MOTA by 3.05. Ray Conditioning then improves MODA by 1.33 and Recall by 1.54. The precision reduction of 0.19 is small relative to the recall gain. BTR leaves detections unchanged, but raises IDF1 by 2.83 and MOTA by 3.43 over VGF + Ray. Thus, VGF and rays modify spatial evidence, whereas BTR changes only temporal association.

\begin{table}[t]
\centering
\caption{Component ablation on WildTrack C1+C3. BTR does not change the detector output.}
\label{tab_component}
\resizebox{\columnwidth}{!}{%
\begin{tabular}{lcccccccccc}
\toprule
Configuration & VGF & Ray & BTR & MODA & MODP & Prec. & Recall & IDF1 & MOTA & MOTP\\
\midrule
TrackTacular & -- & -- & -- & 82.53 & 69.53 & 97.93 & 84.31 & 83.81 & 83.54 & 79.73\\
\midrule
VGF & \checkmark & -- & -- & 85.05 & 74.34 & \textbf{97.95} & 86.87 & 91.24 & 86.59 & 82.35\\
VGF + Ray & \checkmark & \checkmark & -- & \textbf{86.38} & \textbf{76.86} & 97.76 & \textbf{88.41} & 91.38 & 87.64 & 84.33\\
\textbf{GRACE} & \checkmark & \checkmark & \checkmark & \textbf{86.38} & \textbf{76.86} & 97.76 & \textbf{88.41} & \textbf{94.21} & \textbf{91.07} & \textbf{84.37}\\
\bottomrule
\end{tabular}}
\end{table}

The camera-count results also delimit where recovery is useful. At seven cameras, BTR changes MOTA from 90.73 to 91.04, compared with the 3.43-point gain at two cameras. The small seven-camera change is expected because overlapping views already reduce short detection gaps.

\subsection{Effect of Ray Conditioning}

\begin{figure}[!t]
\centering
\begin{minipage}{0.4\columnwidth}
    \centering
    \includegraphics[width=\linewidth]{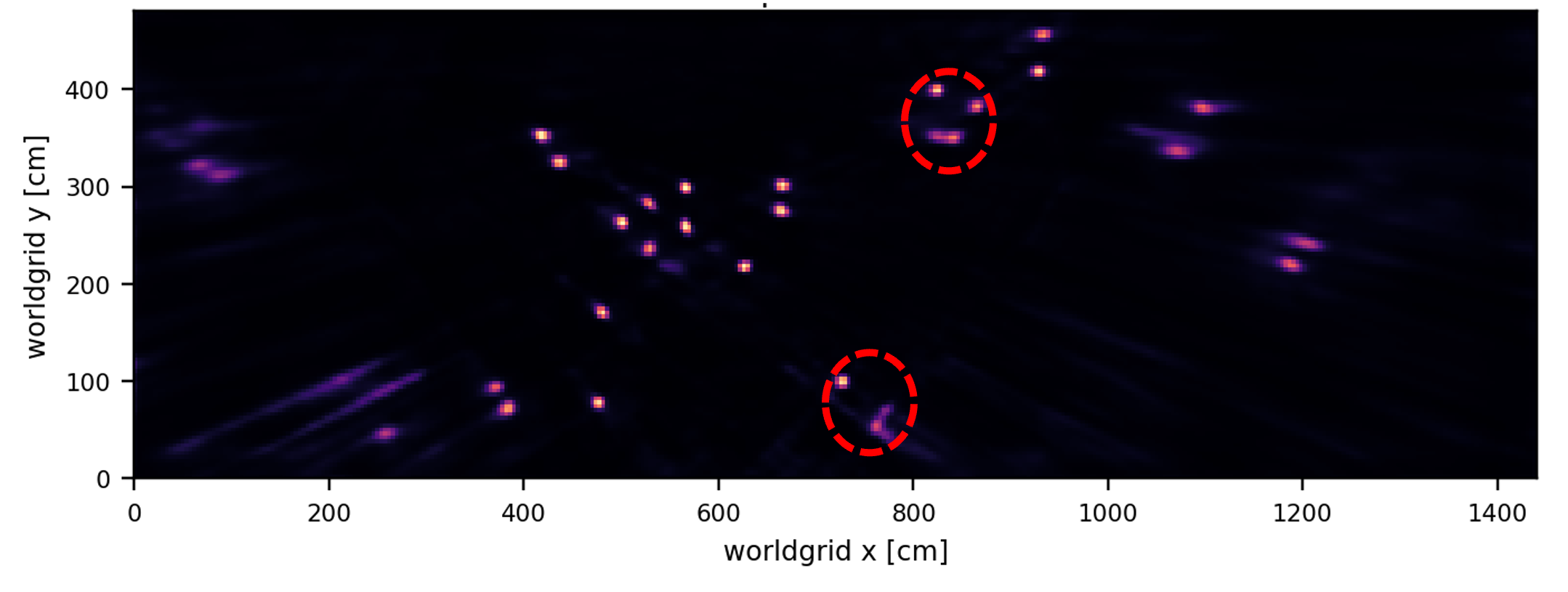}\\[-1mm]
    {\small (a) TrackTacular}
\end{minipage}
\hfill
\begin{minipage}{0.4\columnwidth}
    \centering
    \includegraphics[width=\linewidth]{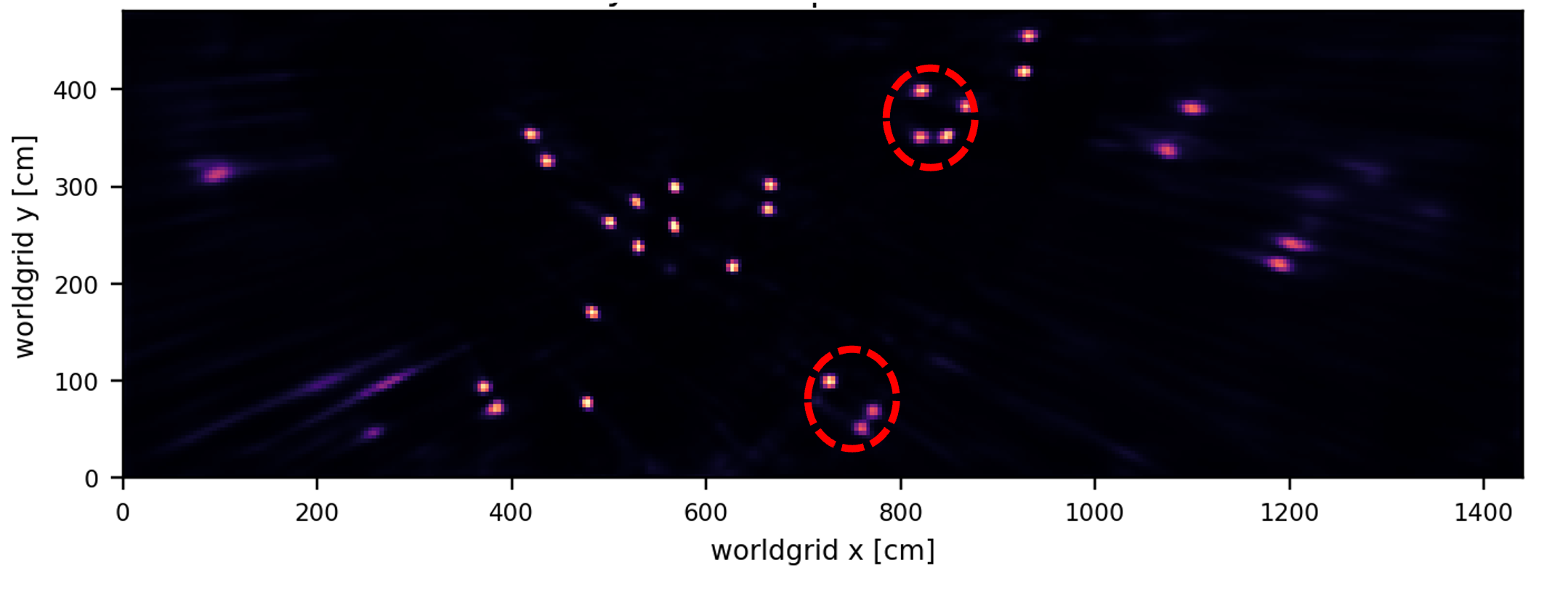}\\[-1mm]
    {\small (b) GRACE}
\end{minipage}
\caption{BEV center heatmaps obtained with the same two WildTrack cameras. TrackTacular shows elongated activations around several pedestrians, while GRACE produces more compact activations. Red circles highlight representative differences.}
\label{fig_rc_effect}
\end{figure}

Ray Conditioning reduces directional spreading in the two-camera setting. Figure~\ref{fig_rc_effect} compares the BEV center heatmaps produced by TrackTacular and GRACE using the same two WildTrack cameras. TrackTacular shows elongated activation patterns around several pedestrians, particularly along the camera viewing direction. GRACE produces more compact activations at the same locations, as highlighted by the red circles.

\FloatBarrier
\section{Conclusion}



GRACE addresses two failures that become more severe when only a few cameras are available: projection errors and short detection gaps. VGF and Ray Conditioning reduce projection errors using camera calibration, while BTR keeps an existing track when the detection score temporarily falls below the normal threshold. With only two WildTrack cameras, GRACE reaches 91.07 MOTA, 1.36 points higher than TrackTacular with seven cameras. On MultiviewX, GRACE improves MOTA by 11.09 points over TrackTacular using the same two cameras.

\clearpage
\begingroup
\interlinepenalty=10000
\bibliographystyle{IEEEbib}
\bibliography{strings,refs}

@inproceedings{hou2020mvdet,
  author = {Hou, Yunzhong and Zheng, Liang and Gould, Stephen},
  title = {Multiview Detection with Feature Perspective Transformation},
  booktitle = {European Conference on Computer Vision},
  year = {2020}
}

@inproceedings{teepe2024earlybird,
  author = {Teepe, Torben and Wolters, Philipp and Gilg, Johannes and Herzog, Fabian and Rigoll, Gerhard},
  title = {EarlyBird: Early-Fusion for Multi-View Tracking in the Bird's Eye View},
  booktitle = {IEEE/CVF Winter Conference on Applications of Computer Vision Workshops},
  pages = {102--111},
  year = {2024}
}

@inproceedings{teepe2024lifting,
  author = {Teepe, Torben and Wolters, Philipp and Gilg, Johannes and Herzog, Fabian and Rigoll, Gerhard},
  title = {Lifting Multi-View Detection and Tracking to the Bird's Eye View},
  booktitle = {IEEE/CVF Conference on Computer Vision and Pattern Recognition Workshops},
  pages = {667--676},
  year = {2024}
}

@inproceedings{chavdarova2018wildtrack,
  author = {Chavdarova, Tatiana and Baqu{\'e}, Pierre and Bouquet, St{\'e}phane and Maksai, Andrii and Jose, Cijo and Bagautdinov, Timur and Lettry, Louis and Fua, Pascal and Van Gool, Luc and Fleuret, Fran{\c{c}}ois},
  title = {WILDTRACK: A Multi-Camera HD Dataset for Dense Unscripted Pedestrian Detection},
  booktitle = {IEEE Conference on Computer Vision and Pattern Recognition},
  pages = {5030--5039},
  year = {2018},
  doi = {10.1109/CVPR.2018.00528}
}

@inproceedings{liu2022petr,
  author = {Liu, Yingfei and Wang, Tiancai and Zhang, Xiangyu and Sun, Jian},
  title = {PETR: Position Embedding Transformation for Multi-View 3D Object Detection},
  booktitle = {European Conference on Computer Vision},
  year = {2022}
}

@inproceedings{xiong2023cape,
  author = {Xiong, Kaixin and Gong, Shi and Ye, Xiaoqing and Tan, Xiao and Wan, Ji and Ding, Errui and Wang, Jingdong and Bai, Xiang},
  title = {CAPE: Camera View Position Embedding for Multi-View 3D Object Detection},
  booktitle = {IEEE/CVF Conference on Computer Vision and Pattern Recognition},
  year = {2023}
}

@inproceedings{shu2023p3d,
  author = {Shu, Changyong and Deng, Jiajun and Yu, Fisher and Liu, Yifan},
  title = {3DPPE: 3D Point Positional Encoding for Transformer-based Multi-Camera 3D Object Detection},
  booktitle = {IEEE/CVF International Conference on Computer Vision},
  pages = {3580--3589},
  year = {2023}
}

@inproceedings{chen2023vedet,
  author = {Chen, Dian and Li, Jie and Guizilini, Vitor and Ambrus, Rares and Gaidon, Adrien},
  title = {Viewpoint Equivariance for Multi-View 3D Object Detection},
  booktitle = {IEEE/CVF Conference on Computer Vision and Pattern Recognition},
  year = {2023}
}

@article{chu2024rayformer,
  author = {Chu, Xiaomeng and Deng, Jiajun and You, Guoliang and Duan, Yifan and Li, Yao and Zhang, Yanyong},
  title = {RayFormer: Improving Query-Based Multi-Camera 3D Object Detection via Ray-Centric Strategies},
  journal = {arXiv preprint arXiv:2407.14923},
  year = {2024}
}

@inproceedings{liu2024raydn,
  author = {Liu, Feng and Huang, Tengteng and Zhang, Qianjing and Yao, Haotian and Zhang, Chi and Wan, Fang and Ye, Qixiang and Zhou, Yanzhao},
  title = {Ray Denoising: Depth-Aware Hard Negative Sampling for Multi-View 3D Object Detection},
  booktitle = {European Conference on Computer Vision},
  year = {2024}
}

@inproceedings{daryani2025camuvid,
  author = {Daryani, Amir Etefaghi and Bhutta, M. Usman Maqbool and Hernandez, Byron and Medeiros, Henry},
  title = {{CaMuViD}: Calibration-Free Multi-View Detection},
  booktitle = {IEEE/CVF Conference on Computer Vision and Pattern Recognition},
  pages = {1220--1229},
  year = {2025},
  doi = {10.1109/CVPR52734.2025.00122}
}

@article{li2026rmcs,
  author = {Li, He and Gui, Jiajia and Kong, Weihang and Zhang, Xingchen},
  title = {Multi-View Pedestrian Detection via Residual Mask Fusion and Cosine Similarity-Based Passive Sampler for Video Surveillance Systems},
  journal = {Future Generation Computer Systems},
  volume = {181},
  pages = {108384},
  year = {2026},
  month = aug,
  doi = {10.1016/j.future.2026.108384}
}

@inproceedings{zhang2022bytetrack,
  author = {Zhang, Yifu and Sun, Peize and Jiang, Yi and Yu, Dongdong and Weng, Fucheng and Yuan, Zehuan and Luo, Ping and Liu, Wenyu and Wang, Xinggang},
  title = {ByteTrack: Multi-Object Tracking by Associating Every Detection Box},
  booktitle = {European Conference on Computer Vision},
  year = {2022}
}

@inproceedings{hyun2023sgt,
  author = {Hyun, Jeongseok and Kang, Myunggu and Wee, Dongyoon and Yeung, Dit-Yan},
  title = {Detection Recovery in Online Multi-Object Tracking With Sparse Graph Tracker},
  booktitle = {IEEE/CVF Winter Conference on Applications of Computer Vision},
  pages = {4850--4859},
  year = {2023}
}

@inproceedings{zhou2020centertrack,
  author = {Zhou, Xingyi and Koltun, Vladlen and Kr{\"a}henb{\"u}hl, Philipp},
  title = {Tracking Objects as Points},
  booktitle = {European Conference on Computer Vision},
  year = {2020}
}

@inproceedings{he2016resnet,
  author = {He, Kaiming and Zhang, Xiangyu and Ren, Shaoqing and Sun, Jian},
  title = {Deep Residual Learning for Image Recognition},
  booktitle = {IEEE Conference on Computer Vision and Pattern Recognition},
  year = {2016}
}

@article{nascimento2006detection,
  author = {Nascimento, Jacinto C. and Marques, Jorge S.},
  title = {Performance Evaluation of Object Detection Algorithms for Video Surveillance},
  journal = {IEEE Transactions on Multimedia},
  volume = {8},
  number = {4},
  pages = {761--773},
  year = {2006},
  doi = {10.1109/TMM.2006.876287}
}

@article{bernardin2008clear,
  author = {Bernardin, Keni and Stiefelhagen, Rainer},
  title = {Evaluating Multiple Object Tracking Performance: The CLEAR MOT Metrics},
  journal = {EURASIP Journal on Image and Video Processing},
  volume = {2008},
  pages = {1--10},
  year = {2008},
  doi = {10.1155/2008/246309}
}

@inproceedings{ristani2016id,
  author = {Ristani, Ergys and Solera, Francesco and Zou, Roger S. and Cucchiara, Rita and Tomasi, Carlo},
  title = {Performance Measures and a Data Set for Multi-Target, Multi-Camera Tracking},
  booktitle = {European Conference on Computer Vision Workshops},
  pages = {17--35},
  year = {2016},
  doi = {10.1007/978-3-319-48881-3_2}
}
\endgroup
\end{document}